\documentclass[12pt, reqno]{amsart}
\usepackage{epsfig}
\usepackage{enumitem}
\def\footnoterule{\kern-3pt \hrule width 1.9 cm \kern2.6pt}
\begin{document}
\centerline{\bf \small{Learning differential equations from data}} 
\vskip 0.4 cm
\centerline{\scriptsize{K. D. Olumoyin}} 
\vskip 0.4 cm
\begin{center}
\begin{minipage}[b]{9.0cm}
\scriptsize{
\noindent 
ABSTRACT. Differential equations are used to model problems that originate in disciplines such as physics, biology, chemistry, and engineering. In recent times, due to the abundance of data, there is an active search for data-driven methods to learn Differential equation models from data. However, many numerical methods often fall short. Advancements in neural networks and deep learning, have motivated a shift towards data-driven deep learning methods of learning differential equations from data. In this work, we propose a forward-Euler based neural network model and test its performance by learning ODEs such as the FitzHugh-Nagumo equations from data using different number of hidden layers and different neural network width. 
}
\end{minipage}
\end{center}
\par
\par
\vskip 0.5 cm
{\small{
\centerline{1. INTRODUCTION}}
}
\vskip 0.3 cm
\noindent 
In the late 1500s, Danish astronomer, Tycho Brahe used geometry and the astronomical tools of his time to collect accurate data about planetary positions. Later, his student Johannes Kepler used Brahe's records to formulate the laws of planetary motion \cite{kepler2002}. This was one of the earliest recorded use of a data-driven approach to formulate and discover the governing laws of our physical and celestial world. On the other hand, we have the fascinating discovery of the heat equation by Mathematician Joseph Fourier \cite{fourier1822}, his was a triumph of intuition and reasoning. And for many decades, such reasoning steered our knowledge of modeling with differential equations.

In recent years, we are seeing a resurgence of data-driven mathematical modeling. We are no longer turning to first principles but increasingly to data-driven approaches to understanding the patterns in complex systems. These systems have the potential to help us understand finance, epidemiology, biology, physics, chemistry, and many other fields. These systems are usually nonlinear and high-dimensional and with applications of mathematical methods in combination with the enormous data available and the improved computational resources, we can make more accurate predictions, forecasts, and extrapolations. 

We are interested in the dynamical systems that describes observed dynamics of biochemical nature such as the Fitzhugh-Nagumo models \cite{FitzHugh1921,Nagumo1962}. In recent years, there is a shift in the modeling approach from first principle to data-driven approaches. This is a merge of data usually big data and deep learning in the analysis and understanding of dynamical systems. This gives us an opportunity to validate assumptions about underlying physical and biological laws that govern these observed dynamics.
\vskip 0.5 cm
{\small{
\centerline{2. PRELIMINARY}}
}
\vskip 0.3 cm
\noindent
Advancement in machine learning and data science have made it possible to learn complex patterns from data. Data-driven approaches including neural networks and nonlinear regression have been applied to dynamical systems~\cite{sindy2016,zabaras2020,ren2021}. 
Data-driven approaches are not replacing good intuition and reasoning but instead are unlocking new possibilities~\cite{olumoyin2021}. In~\cite{Mamikon2018}, the authors used Gaussian processes to discover differential equations from data. The limitation with their approach is that they can only learn linear systems. An advancement was made in~\cite{sindy2016}, where the authors formulated a sparse identification of a nonlinear dynamical system for extracting governing equations from data. Their approach is able to learn nonlinear systems however it can easy become computationally expensive when scaled to higher dimensions. It requires some knowledge of the dynamics while constructing the library of functions that aid the neural network to learn the dynamics. One of the most successful approaches in the last five years is the physics-informed neural network (PINN) introduced in~\cite{Raissi2019}, where the form of the differential equation is assumed to be known and the task is to learn the parameters of the differential equation from data.PINN excels when trained with sparse data because it uses the governing equations as part of the training process. PINN is generally limited to low-dimensional parameter identification. It has not performed well in modeling differential equations with sharp gradients, and it has become necessary to use collocation points in the differential equation residuals. This results in a huge computational cost in training. However, in a true sense of learning differential equations from data, we desire approaches that do not assume that the dynamics of the differential equation is known. The following works have succeeded in modeling differential equations from data, where the dynamics of the differential equation is not assumed to be known~\cite{zabaras2020} and~\cite{ren2021}. In this work, we propose a forward-Euler feedforward neural network.

\par
\noindent
\vskip 0.5 cm
{\small{
\centerline{3. METHODOLOGY} }}
\vskip 0.3 cm
\noindent
\par			

In this work, we will do the following:

\begin{enumerate}
\item Formulate a neural network,
This network is used to advance the neural network prediction of the differential equation from a temporal state to the next higher temporal state. 

\item Imposition of initial and boundary conditions so that we can have a well-posed optimization problem during the training. 

\item A residual differential equation connection inspired by the forward-Euler scheme will be used to construct the loss function. If we seek to improve the network, we will have to use higher-order temporal schemes to design the loss function. The use of higher-order temporal schemes will require changes in the network architecture. In this work, we want to demonstrate that a loss function built using a forward Euler temporal scheme in the network formulation is capable to learn the form of a differential equation from data to high accuracy.
\end{enumerate}



%
%

%

Our goal is to develop a neural network that can be used in an inverse problem, where data is available. Neural network approaches are suitable for differential equations because they can scale to high dimensions. They can also solve forward and inverse problems with sparse data. As such, the formulation of such a network can be used as a new numerical solver for various time-dependent differential equations given specific initial and boundary conditions. 







In totality, our approach provides an approximation of a discrete solution $\mathbf{u}(x,t;\theta)$ denoted as $\mathbf{u}^{pred}$ that satisfies \eqref{eq1} for specifically given an initial condition. Here $\theta$ denotes the neural network trainable parameters.


\vskip 0.5 cm
{\small{
\centerline{3.1 Data generation} }}
\vskip 0.3 cm


We will learn the FitzHugh-Nagumo system (FHN)~\cite{FitzHugh1921,Nagumo1962} from data. FHN is a simplification of the Hodgkin-Huxley model~\cite{Hodgekin1952}. The motivation for the FHN model was to isolate conceptually the essential mathematical properties of excitation and propagation from the electrochemical properties of sodium and potassium ion flow. The model consists of a voltage-like variable $\it{u}$ having cubic nonlinearity that allows regenerative self-excitation via a positive feedback, and a recovery variable $\it{v}$ having a linear dynamics that provides a slower negative feedback.

The partial differential equation version of FHN is given below:
\begin{equation} \label{fhn_eqn}
\begin{split}
\frac{\partial u}{\partial t}  &= \frac{1}{c}(u-\frac{1}{3}u^3-v) + \triangle u \\
\frac{\partial v}{\partial t}  &= c(u-av+b)
\end{split}
\end{equation}

The parameters in  \eqref{fhn_eqn} satisfies the following inequalities, $0<a<1$, $b>0$, $c>0$. An appropriate initial condition is chosen by finding an homogeneous solution of \eqref{fhn_eqn}, which satisfies the ordinary differential equations below  \eqref{fhn_ode}:

\begin{equation} \label{fhn_ode}
\begin{split}
\frac{d u}{d t}  &= \frac{1}{c}(u-\frac{1}{3}u^3-v) \\
\frac{d v}{d t}  &= c(u-av+b)
\end{split}
\end{equation}

\begin{figure}[!htbp]
\centerline{\includegraphics[width=8.0 cm]{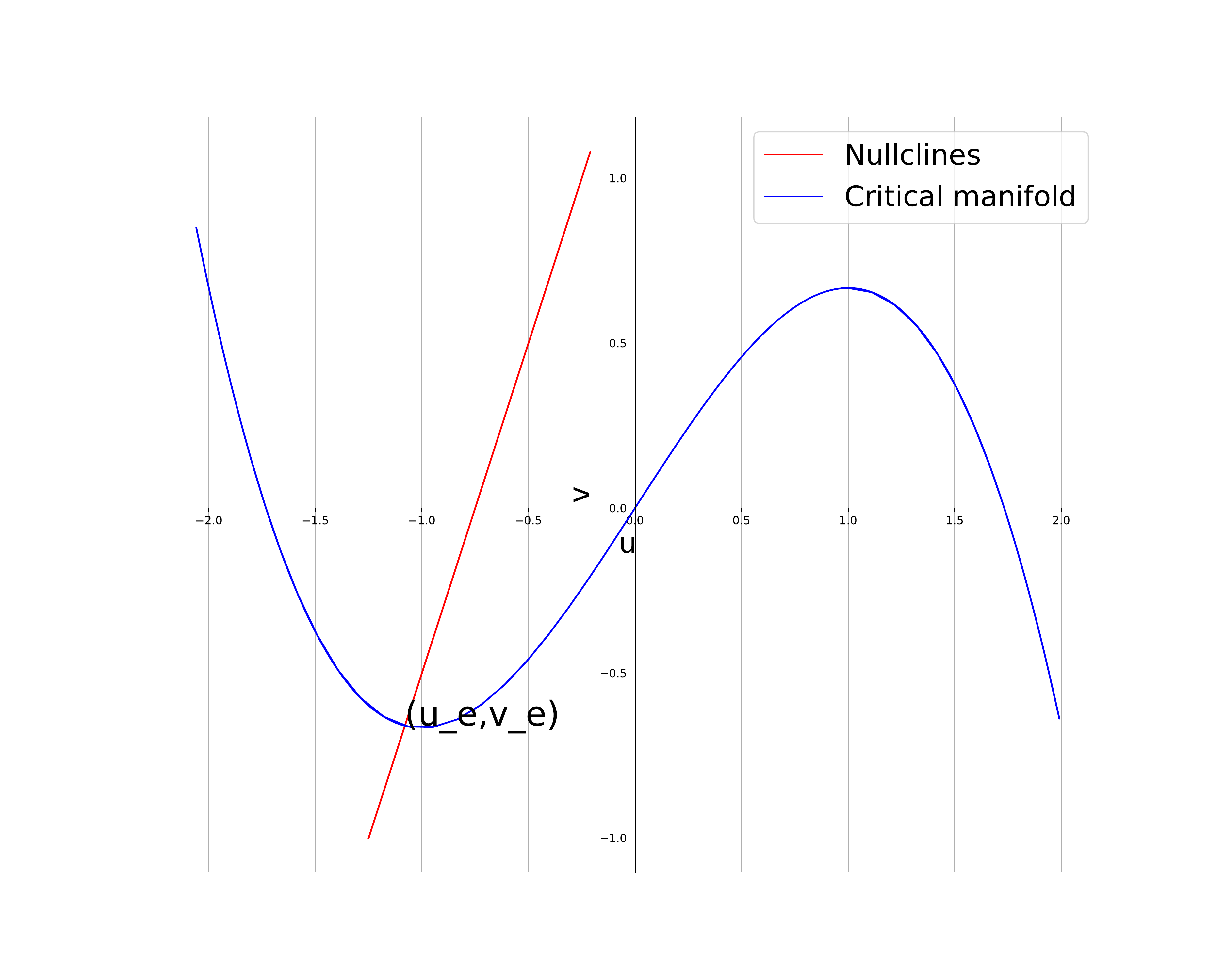}}
\caption{Diagram of the nullcline and critical manifold for homgeneous FitzHugh-Nagumo model.}
\label{nullclines}
\end{figure}

The equilibrium solution of \eqref{fhn_ode} is $(u_e,v_e)$ as shown in Figure~\eqref{nullclines}. Where the nullclines of the FHN equations are a line and a cubic that intersect in a single rest point.

We consider the following initial conditions $u(x,0)$ and $v(x,0)$ for system \eqref{fhn_eqn} given as follows:

\begin{equation*}
u(x,0) = u_0(x), v(x,0) = v_0(x)
\end{equation*}

where,

\begin{equation*}
u_0(x) = \begin{cases}
u_e &\text{if $x\geq\frac{L}{2}$}\\
-u_e &\text{if $x<\frac{L}{2}$}
\end{cases}
\end{equation*}

\begin{equation*}
v_0(x) = -v_e
\end{equation*}

and $x \in [0,L]$.

Training and testing dataset is generated by solving \eqref{fhn_ode} using ode solvers such as scipy's odeint function.

\vskip 0.5 cm
{\small{
\centerline{3.2 A discrete physics loss function based Neural Network} }}
\vskip 0.3 cm

We consider the general form of a multi-dimensional, nonlinear, coupled PDE system of the form

\begin{equation}\label{eq1}
    \mathbf{u}_t(x,t) = \mathcal{F}[\mathbf{u},\mathbf{u}^2, \ldots, \mathbf{u}_x, \mathbf{u}_{xx}, \ldots; \bf{\lambda} ]
\end{equation}

where $\mathbf{u}(x,t)$ is the physical quantity we want to study. The temporal domain is $t\in[0,T]$ and the spatial domain is $x\in \Omega \subset \mathbf{R}$. $\mathbf{u}_t(x,t)$ is the first-order time derivative term and $\mathbf{u}_x$ denotes the gradient operator with respect to the spatial variable $x$. $\mathcal{F}[\cdot]$ is the nonlinear functional parameterized by $\bf{\lambda}$. Eq \eqref{eq1} satisfies appropriate initial and boundary conditions.

\begin{figure}[htbp]
\centering
\includegraphics[width=10cm]{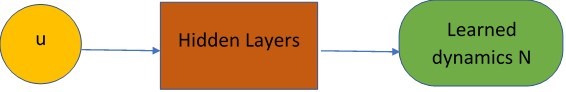}
\caption{Schematic of ForwardEulerNet}
\label{fen_fhn_sch}
\end{figure}

We will learn the FitzHugh-Nagumo system(FHN)~\cite{FitzHugh1921,Nagumo1962} from data. We consider \eqref{fhn_ode} in 1 spatial dimensions, and $t>0$.Here, we assume only a temporal domain $t\in[0,T]$, 
and $\mathbf{u}$ denotes the variables $(u,v)$. In this approach, we do not assume the form of the right hand side of \eqref{fhn_ode}. Suppose $\mathbf{\hat{u}}$ is the target variable for which there is a known measurements, we can formulate a neural network that take as input; $\mathbf{\hat{u}}$, and the temporal domain $t\in[0,T]$. The output of this neural network is the learned right hand side of \eqref{fhn_ode} denoted as $\mathcal{F}$. This neural network is a feedforward neural network, see Figure~\eqref{fen_fhn_sch}.

The loss function can then be formulated as follows

\begin{equation}\label{loss}
\mathcal{R}(t;\theta) = \mathbf{\hat{u}}_t - \mathcal{F}[t,\mathbf{\hat{u}}, \ldots; \bf{\lambda}, \bf{\theta} ]
\end{equation}

A discrete loss is used in training the neural network, where the discrete $\mathcal{F}$ is given by $\mathcal{N}$.
The discrete loss function is inspired by the forward Euler method to connect the temporal state $\mathbf{\hat{u}}_{i}$ and  temporal state $\mathbf{\hat{u}}_{i+1}$. 

\begin{equation}
\mathbf{\hat{u}}_{i+1} = \mathbf{\hat{u}}_{i} + \partial t \cdot \mathcal{N}[\mathbf{\hat{u}};\theta], \quad \partial t =\frac{T}{K}, K \in \mathbf{R}^{+}
\end{equation}

The trainable parameters $\theta$ are optimized during training of the forward-Euler based neural network that  minimizes the mean squared error loss function (MSE) given as follows

\begin{equation}
MSE = \sum_{i=1}^{K}|-\mathbf{\hat{u}}_{i+1} + \mathbf{\hat{u}}_{i} + \partial t \cdot \mathcal{N}[\mathbf{\hat{u}};\theta]|^2, \quad \partial t =\frac{T}{K}, K \in \mathbf{R}^{+}
\end{equation}

When training is done, we obtain $\mathcal{N}$, which is the learned dynamics of an ODE system such as eq \eqref{fhn_ode}. Now to obtain learned $\mathbf{u}$ denoted as $\mathbf{u}^{pred}$, we use an ODE solver such. Using a solver such as the scipy odeint. We observe that the learned model parameters $\lambda=[a,b,c]$ in eq~\eqref{fhn_ode} are embedded into the dynamics $\mathcal{N}$.

\[
\mathbf{u}^{pred} = odeint(\mathcal{N}, (u(0),v(0)), tspan)
\]

\begin{figure}[htbp]
\includegraphics[width=13cm]{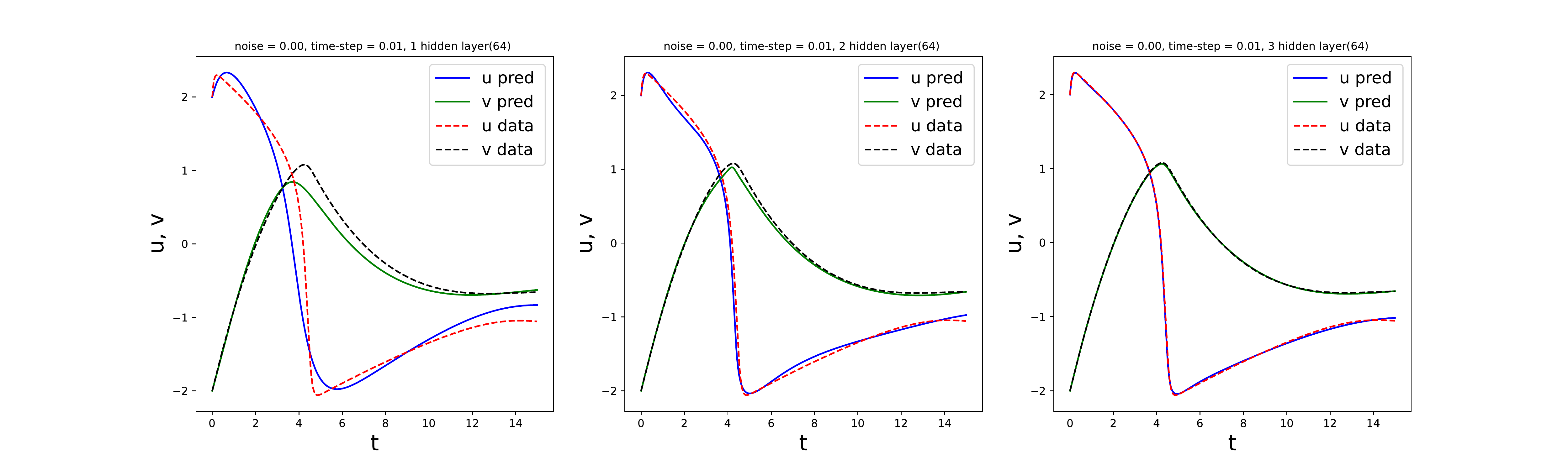}
\caption{No noise, $\triangle t$ = 0.01, 64 neurons }
\label{fhn_ode1}
\end{figure}

\begin{figure}[htbp]
\includegraphics[width=13cm]{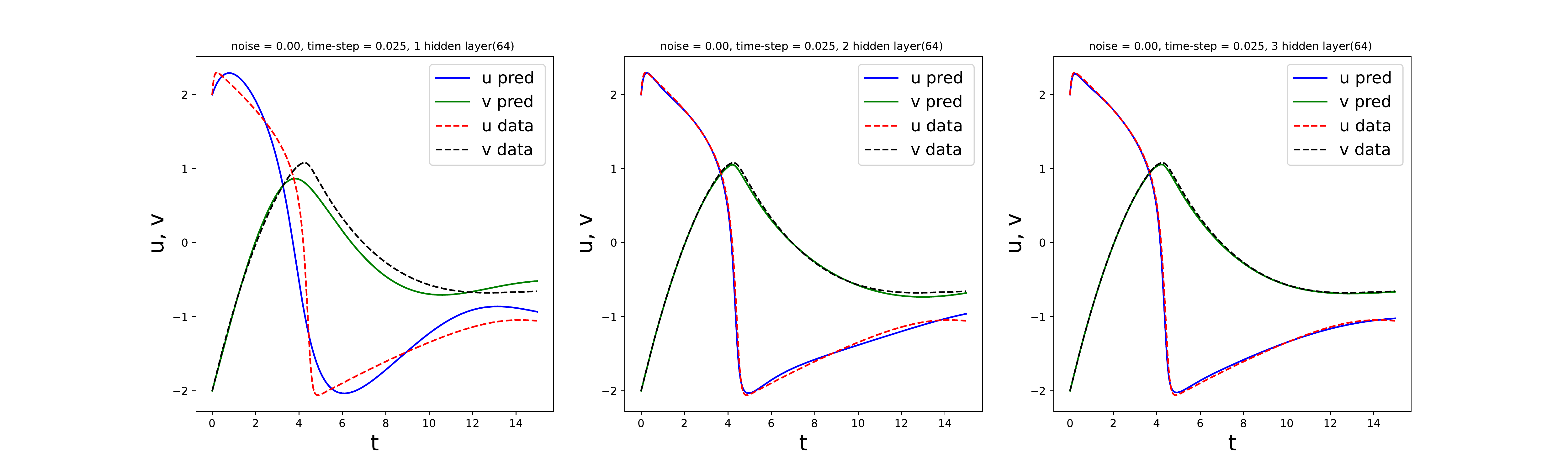}
\caption{No noise, $\triangle t$ = 0.025, 64 neurons }
\label{fhn_ode2}
\end{figure}

\begin{figure}[htbp]
\includegraphics[width=13cm]{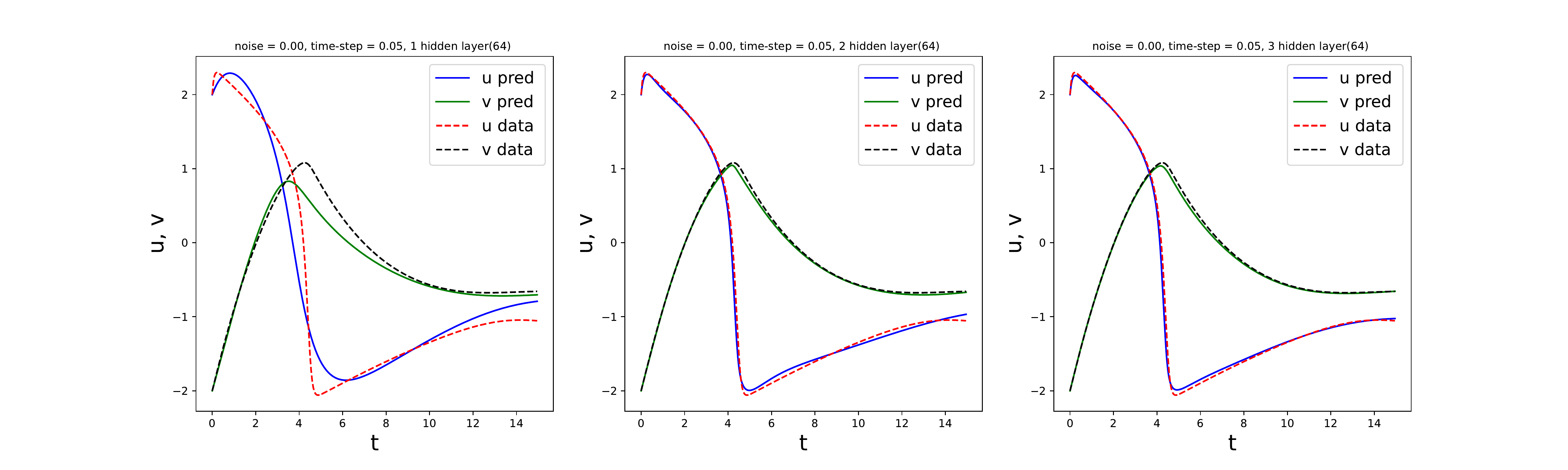}
\caption{No noise, $\triangle t$ = 0.05, 64 neurons }
\label{fhn_ode3}
\end{figure}

\begin{figure}[htbp]
\includegraphics[width=13cm]{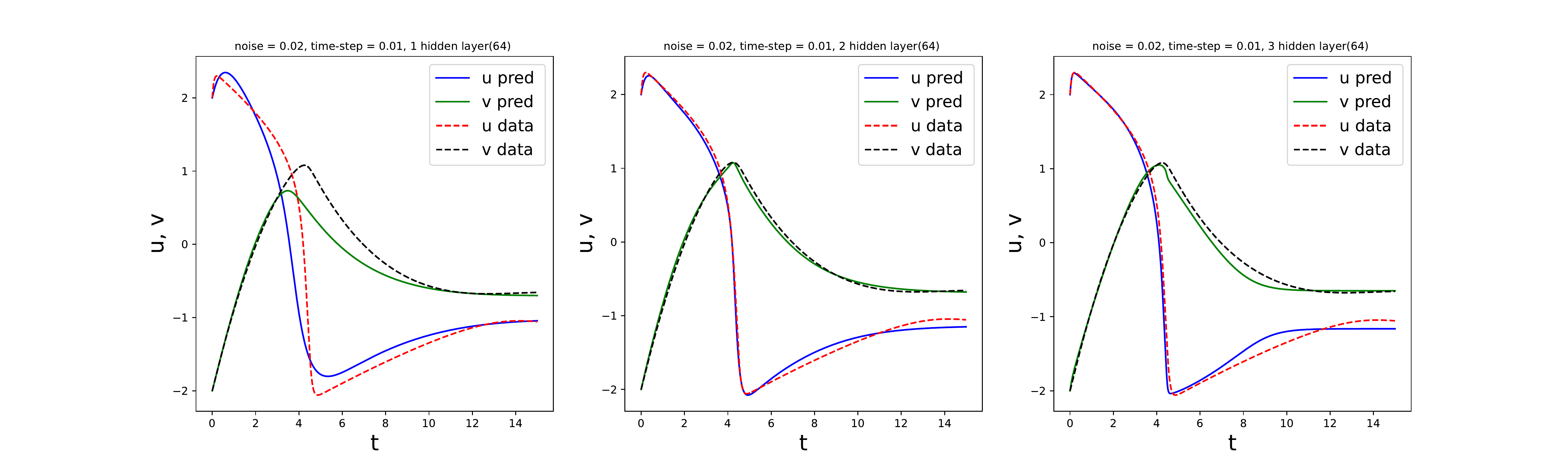}
\caption{$2 \%$ noise, $\triangle t$ = 0.01, 64 neurons }
\label{fhn_ode4}
\end{figure}

\begin{figure}[htbp]
\includegraphics[width=13cm]{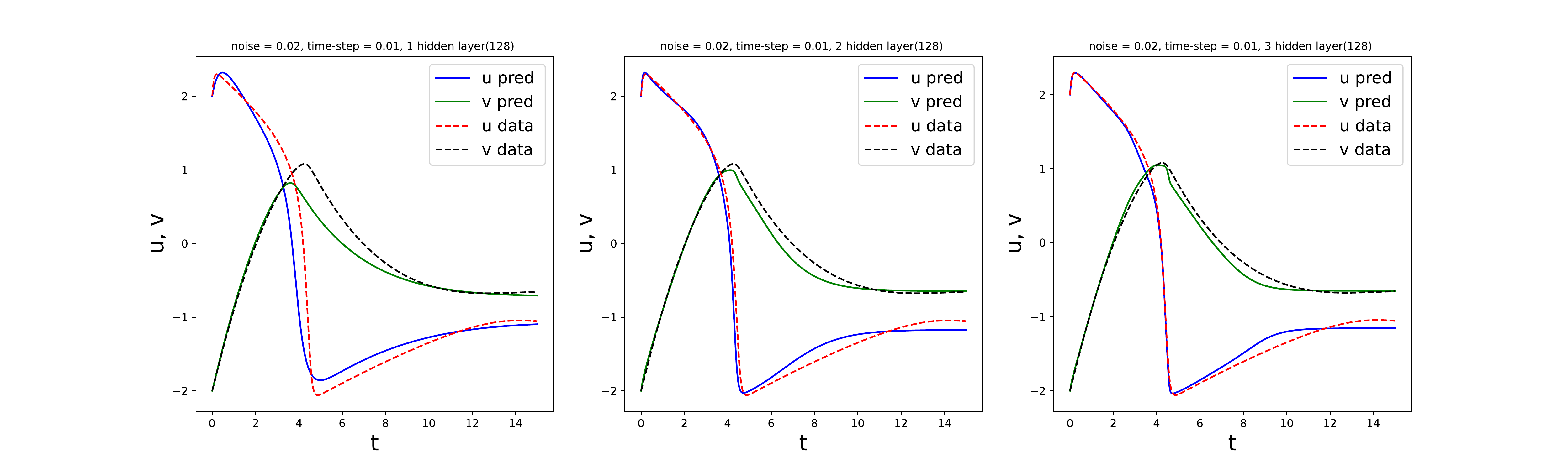}
\caption{$2 \%$ noise, $\triangle t$ = 0.01, 128 neurons }
\label{fhn_ode5}
\end{figure}

\begin{figure}[htbp]
\includegraphics[width=13cm]{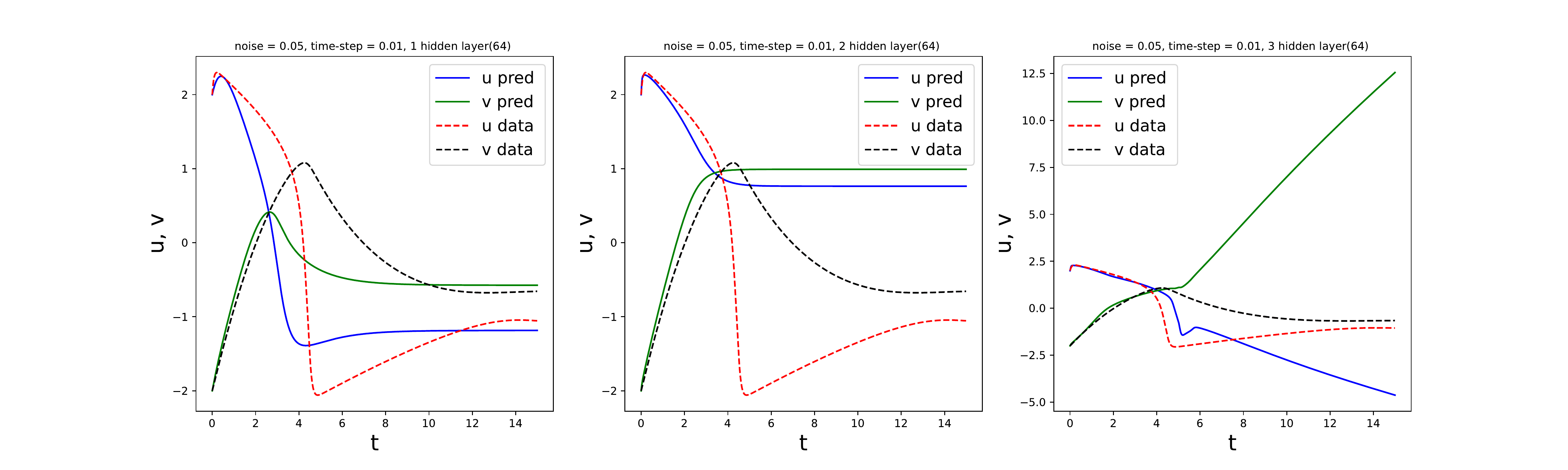}
\caption{$5 \%$ noise, $\triangle t$ = 0.01, 64 neurons }
\label{fhn_ode6}
\end{figure}

In Figures~\eqref{fhn_ode1}--\eqref{fhn_ode6}, we obtained $\mathbf{u}^{pred}$ and compared with the target $\mathbf{\hat{u}}$.
In Figure~\eqref{fhn_ode6}, we observe that when we add $5\%$ noise to the training data, we are no longer able to learn an accurate dynamics of ~\eqref{fhn_ode} from data.

\vspace{0.3 cm}
\par
{\small{
\centerline{4. CONCLUDING REMARKS}}}
\vskip 0.4 cm
\par
\noindent 

We present results and performance of the forward-Euler neural network approach that learns the FitzHugh-Nagumo ODE~\eqref{fhn_ode} from data. 
In Table~\ref{table_f}, we observe that increasing the number of hidden layers, the MSE error of the variable $u$ produces slightly more accurate results when the width (number of neurons per layer) of the neural network is 32, there are not much difference in the MSE error when we increase the hidden layers and the width is 64, 128, and 256. However, the benefit of increasing the width is more noticeable in the case of one hidden layer.

\begin{table}[!htbp]
\begin{tabular}{ |p{2.5cm}|p{1.4cm}|p{1.4cm}|p{1.4cm}|p{1.4cm}|  }
\hline
\multicolumn{5}{|c|}{Width} \\
\hline
Hidden layers & 32 & 64 & 128 & 256 \\
\hline
1 & 7.6e-03 & 4.3e-03 & 1.3e-03 & 2.3e-03 \\
2 & 4.8e-03 & 3.0e-03 & 2.3e-03 & 3.53-03 \\
3 & 4.9e-03 & 4.4e-03 & 2.8e-03 & 3.3e-03 \\
\hline
\end{tabular}
\caption{Table showing MSE error of the variable $u$, where we assumed a noise free training data and set the time step to $0.01$.}
\label{table_f}
\end{table}

\par
\vskip 0.5 cm
\vskip 0.3 cm
{\small{
\centerline{REFERENCES}
}}
\scriptsize{

}
\par
\noindent 
\tiny{DEPARTMENT OF MATHEMATICS,
Middle Tennessee State University, Murfreesboro, TN, USA
\par 
\noindent 
{\it E-mail address}: {\tt kayode.olumoyin@mtsu.edu} 
\par
}
\end{document}